\documentclass[10pt,twocolumn,letterpaper]{article}

\usepackage{iccv}
\usepackage{times}
\usepackage{epsfig}
\usepackage{graphicx}
\usepackage{amsmath}
\usepackage{amssymb}
\usepackage{authblk}
\makeatletter
\renewcommand\AB@affilsepx{     \protect\Affilfont}
\makeatother

\usepackage[breaklinks=true,bookmarks=false]{hyperref}

\iccvfinalcopy 

\DeclareMathOperator*{\argmin}{arg\,min}

\def\ie{\emph{i.e.~}}

\ificcvfinal\fi
\begin{document}

\title{Dual Grid Net: hand mesh vertex regression from single depth maps}

\author[1]{Chengde Wan}
\author[1]{Thomas Probst}
\author[1,3]{Luc Van Gool}
\author[2]{Angela Yao}
\affil[1]{ETH Z\"urich}
\affil[2]{National University of Singapore}
\affil[3]{KU Leuven}

\maketitle

\begin{abstract}
   We present a method for recovering the dense 3D surface of the hand by regressing the vertex coordinates of a mesh model from a single depth map. To this end, we use a two-stage 2D fully convolutional network architecture. 
   In the first stage, the network estimates a dense correspondence field for every pixel on the depth map or image grid to the mesh grid. In the second stage, we design a differentiable operator to map features learned from the previous stage and regress a 3D coordinate map on the mesh grid. Finally, we sample from the mesh grid to recover the mesh vertices, and fit it an articulated template mesh in closed form.
   
   During inference, the network can predict all the mesh vertices, transformation matrices for every joint and the joint coordinates in a single forward pass.  
   When given supervision on the sparse key-point coordinates, our method achieves state-of-the-art accuracy on NYU dataset for key point localization while recovering mesh vertices and a dense correspondence map. Our framework can also be learned through self-supervision by minimizing a set of data fitting and kinematic prior terms.  With multi-camera rig during training to resolve self-occlusion, it can perform competitively with strongly supervised methods without any human annotation.
\end{abstract}

\section{Introduction}
\begin{figure}
\label{fig:intro}
\includegraphics[width=\linewidth]{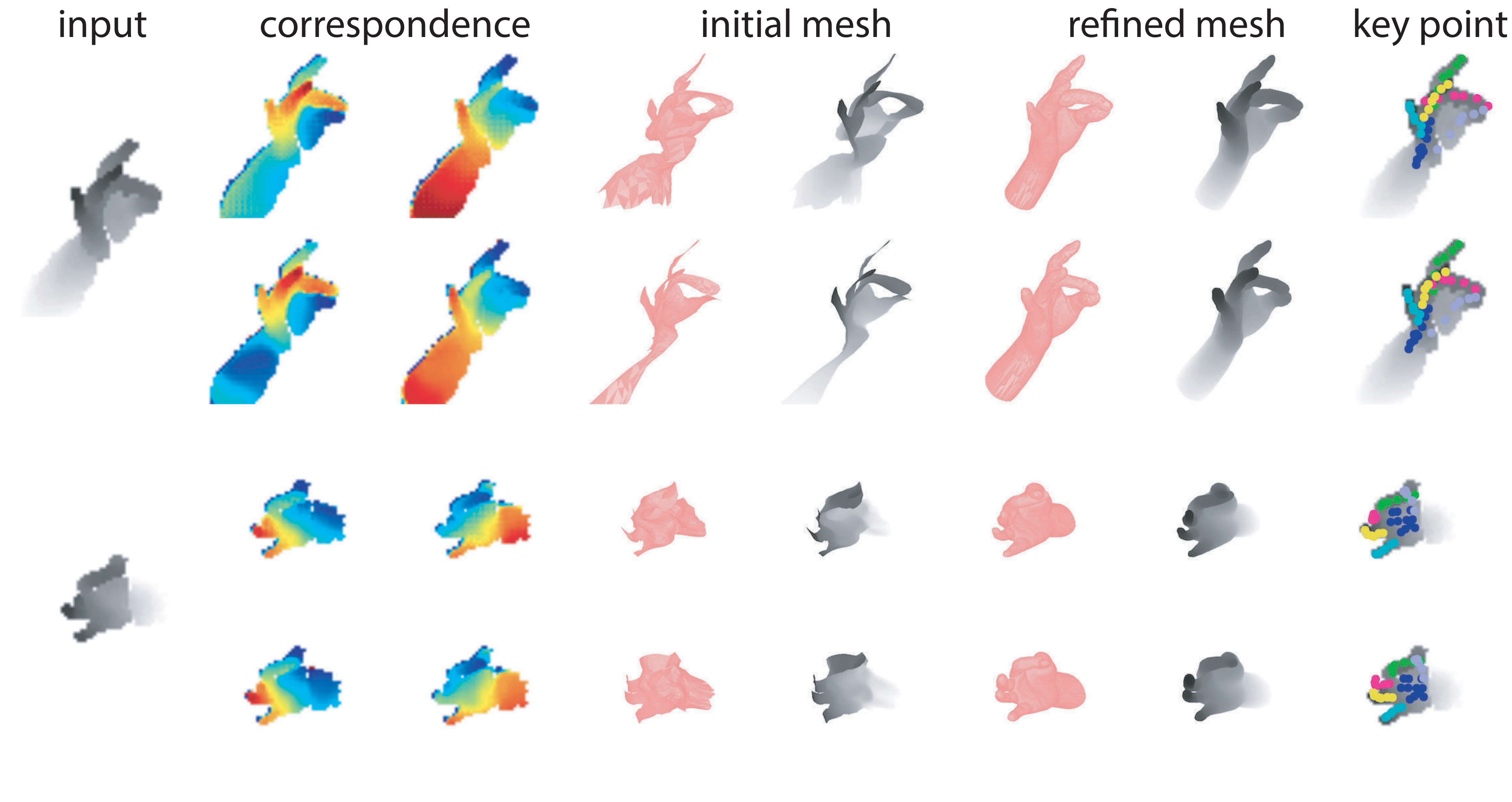}
\caption{\textbf{Qualitative Results.} In each group, upper rows are results supervised with key-point annotation and lower rows are self-supervision result without any human label. We visualize the correspondence map with each mesh coordinate, the rendered shading and depth map of the initial estimated mesh model and refined ones, as well as key-point. More qualitative results will be shown in supplementary material.}
\end{figure}

We consider the problem of estimating the 3D object shape and pose from single depth images.  Specifically, we are interested in estimating the surface mesh vertices of the human hand model from depth maps.
Compared to skeleton joints, dense mesh vertices provide both pose and shape information of the hand and enable a much wider scope of virtual and mixed reality applications. For example, one can directly pose the virtual hand in a VR game, or overlay a user's hand surface with another texture map in mixed reality. Furthermore, the modelling of surface contacts when manipulating virtual objects on screen can be improved with mesh representations.

The estimation of mesh vertices as opposed to skeleton joints is significantly more challenging in several regards.  First, the scale of the problem increases by several magnitudes, \ie to reasonably represent a human hand, one needs thousands of mesh vertices, as opposed to tens of joint positions and angles in a standard skeleton model. Secondly, getting accurate 3D ground truth for the thousands of vertices from real-world data is extremely difficult, even though having large amounts of labelled training data is crucial for data-driven learning based methods. 

Several recent advances have been made to estimate mesh vertices with deep learning, including the use of voxel net~\cite{varol2018bodynet}, graph convolutions~\cite{ranjan2018generating, ge2019handshapepose}, and directly estimating shape parameters and joint angles~\cite{boukhayma20193d, zhang2019end,Malik2018DeepHPSEE}.  These approaches, while having made significant advances for hand pose estimation, have several drawbacks. They tend to be restricted to certain mesh topologies, feature a large number of parameters to learn, and have limited spatial resolutions. 

In this work, we propose to solve the problem of mesh vertex regression with a fully convolutional architecture. Our approach is highly efficient and flexible enough to handle different mesh topologies. Moreover, we can also capture very fine spatial detailing through per-pixel correspondences to a mesh model, thereby allowing for better alignment between the mesh model and depth observations.  

We parameterize the mesh vertices with a 2D embedding; the embedding vector associated with each vertex is its ``intrinsic'' property, \ie only related to its location on the hand mesh surface, regardless of the hand pose, shape and camera view point. In turn, the 3D coordinate of the mesh vertex is considered ``extrinsic''. Similar to digital imaging, we discretize the embedding space by placing a 2D grid, namely the \emph{mesh grid} on the mesh embedding. Both ``intrinsic'' and ``extrinsic'' properties for each mesh vertex can be approximated in terms of a weighted sum with properties of its neighbour points on the grid.

At the core of our method are two 2D fully convolutional networks, applied to the image and mesh estimates consecutively (see overview in Figure~\ref{fig:framework}).  Linking the networks together is a 2D embedding which makes for an efficient way to propagate errors directly from the irregular representation of a mesh to the regular and ordered representation of an image.  
To refine the estimated mesh, we design a simple kinematic module.  Given a template hand mesh model with one-to-one correspondences to the estimated mesh, we solve for a similarity transform through singular value decomposition (SVD). We then re-pose the template mesh based on the transform, resulting in a denoised mesh surface together with key points.
Since SVD has closed form solutions and is a differentiable operator, one can also place supervision on top of the estimated key points.  

For training our model, we propose a self-supervision scheme that minimizes a geometric model-fitting energy as a training loss. The model's accuracy steadily improves with increasing amounts of data seen, even without any human-provided labels. Finally, since correspondences between observed hand pixels and the mesh are estimated in a differentiable way, we can optimize the correspondences jointly with the disparity between the correspondence pairs during model-fitting. This differs from and complements standard ICP optimization methods.

Our contributions can be summarized as follows,
\begin{itemize}
    \item We propose a new fully convolutional network architecture for regressing thousands of mesh vertices in an end-to-end manner. While our method works with single depth maps, the network architecture is ready to handle RGB image case without any additional changes.
    \item A self-learning scheme is proposed for training the network; without any human labels, our network achieves competitive results when compared to fully supervised state-of-the-art. Such a learning approach offers a new and accurate way of annotating real-world data and thereby solves one of the key difficulties in making progress for hand pose estimation. 
    \item We bridge the gap between data-driven discriminative methods and optimization-based model-fitting and enjoy benefits from both sides: accuracy that improves with the amount of data encountered, while not needing human-provided annotations. 
\end{itemize}

\section{Related Works}
\textbf{Hand pose estimation.} Deep learning has significantly advanced state-of-the-art for hand pose estimation.  The general trend has been the development of ever deeper and more sophisticated neural network architectures~\cite{chen2017pose,oberweger2017deepprior++,chen2018shpr,ge2018point,moon2018v2v,ge2018hand,wan2017dense}. However, such progress has also hinged on the availability of large amounts of annotated data~\cite{tompson2014real,yuan2017bighand2,simon2017hand}. Obtaining accurate annotations, even for simple 3D joint coordinates, is extremely difficult and time consuming.  
Annotations generated by manually initializing trackers~\cite{tompson2014real,oberweger2016efficiently} require carefully designed interfaces for 3D annotation on a 2D screen and there is often little consensus between human annotators~\cite{supancic2015depth}.   
Motion-capture rigs\cite{simon2017hand} and auxiliary sensors\cite{yuan2017bighand2} are fully automatic but are limited in the scenes in which they can be deployed.
To mitigate the limitations of annotation, semi-supervised approaches~\cite{wan2017crossing,cai2018weakly,poier2018learning} and approaches coupling synthesized with real data~\cite{shrivastava2017learning,poier2019murauer,rad2018feature} have also been proposed.

An alternative line of work\cite{taylor2014user,qian2014realtime,tagliasacchi2015robust,sharp2015accurate,tang2015opening,joseph2016fits,Sridhar2016,taylor2017articulated} tackles hand pose estimation by minimizing a model-fitting error. 
Model-fitting needs little to no human labels, but the accuracy is heavily dependent on the careful design of the energy function.
A recent trend tries to bridge the gap between data-driven and model-fitting approaches~\cite{Tung2017,dibra2017refine,ge2019handshapepose} by using a differentiable renderer and incorporating the model-fitting error as a part of the training loss. Our work resembles these methods, though we have two key differences. 
First, we re-parameterize the mesh with a 2D embedding, which allows us to use a 2D fully convolutional network architecture.  Secondly, we can apply self-supervision on both the image grid and the mesh grid, leading to efficient gradient flow during back-propagation.

\textbf{Human mesh model recovery from single image.} Data-driven methods have greatly advanced the field of 3D reconstruction of both shape and pose of the full human body~\cite{taylor2012vitruvian,Wei_2016_CVPR,bogo2016keep,omran2018neural,tan2017indirect,pavlakos2018humanshape,Tung2017,hmrKanazawa17,varol2018bodynet}, face~\cite{joo2018total,lombardi2018deep,Yu2018,ranjan2018generating} and hands~\cite{taylor2017articulated,joo2018total,Malik2018DeepHPSEE,zhang2019end,boukhayma20193d,ge2019handshapepose}. Earlier works were focused on landmark detection\cite{bogo2016keep}, segmentation\cite{taylor2017articulated} and finding correspondences~\cite{taylor2012vitruvian,Wei_2016_CVPR,joo2018total,Yu2018}, and performed a model-based optimization to fit the mesh in a subsequent step.  
However, recent trends have shifted towards end-to-end learning of the mesh with neural networks. For example,~\cite{omran2018neural,hmrKanazawa17,pavlakos2018humanshape,Tung2017,zhang2019end,boukhayma20193d,Malik2018DeepHPSEE} directly estimate shape parameters and joint angles of the mesh.  However, such methods are sensitive to perturbations, since small offsets from only one dimension of the estimation easily propagates to many mesh vertices along the kinematic tree. 
In~\cite{tan2017indirect, lombardi2018deep, ranjan2018generating,ge2019handshapepose}, auto-encoders are used with various decoder structures and outputs, including graph convolution to mesh vertices~\cite{ranjan2018generating,ge2019handshapepose}, VoxelNet to 3D occupancy grids\cite{varol2018bodynet}, and fully connected and transposed convolutions to  silhouette~\cite{tan2017indirect} and texture and mesh vertices~\cite{lombardi2018deep}.  Unlike any of these works, our approach is based on correspondence estimation.  Yet we also differ from other correspondence-based methods~\cite{Wei_2016_CVPR,taylor2012vitruvian,alp2017densereg,joo2018total,Yu2018} in that we estimate mesh vertices with a single forward pass in the framework.

\textbf{3D Network Architectures.} It is highly intuitive to parameterize 3D inputs and or outputs as an occupancy grid or distance field and use for example a 3D voxel net~\cite{ge20173d,varol2018bodynet,moon2018v2v}. However, such an architecture is parameter heavy and severely limited in spatial resolution. PointNet~\cite{qi2016pointnet} is a light-weight alternative and while it can interpret 3D inputs a set of un-ordered points, it also largely ignores spatial contexts which may be important downstream.

Since captured 3D inputs are inherently object surfaces, it is natural to consider them as 
a 2D embedding in 3D Euclidean space.  As such, several works ~\cite{cnn_graph,kostrikov2018surface,ranjan2018generating}
have modeled mesh surfaces as a graph and have applied graph network architectures to capture intrinsic and extrinsic geometric properties of the mesh. Our method also works on the hand surface, but it is a much simpler and more flexible network architecture which is easier to train and can handle different mesh topologies.
Our method most resembles ~\cite{su18splatnet,Matan2018} by mapping high dimension data to a 2D grid. However, instead of just working on points from depth map, we propose a dual grid network architecture, enabling the mapping of heterogeneous data from Euclidean space to mesh surfaces and vice versa.

\section{Dual Grid Net}~\label{sec:network}
\begin{figure*}
    \centering
    \includegraphics[width=\linewidth]{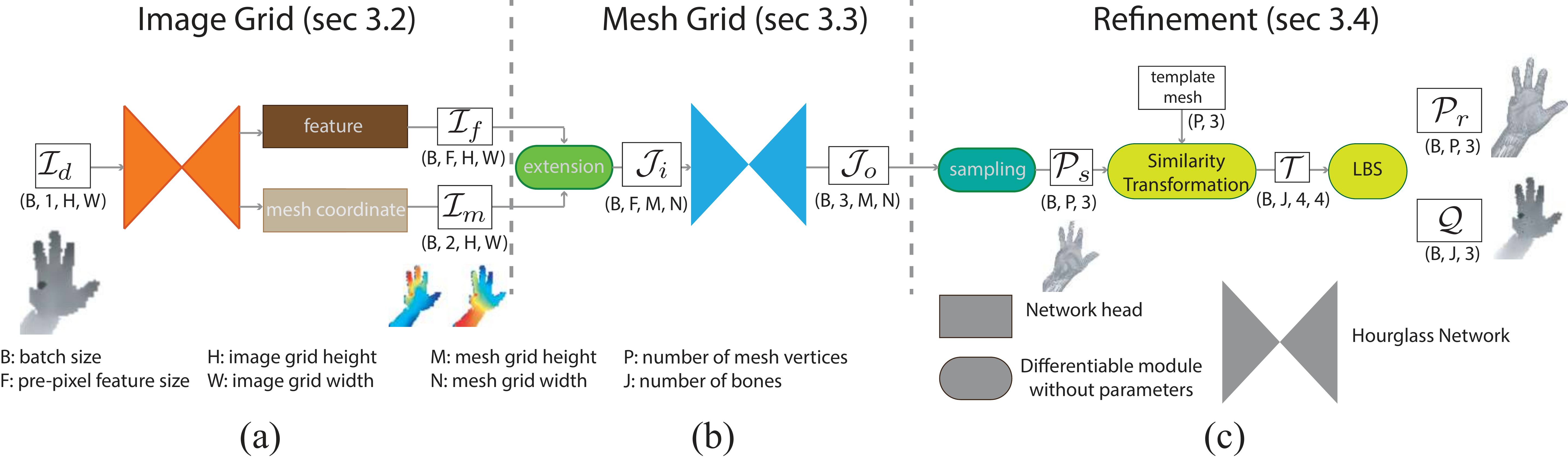}
    \caption{\textbf{System Framework.} Starting from a depth map of the segmented hand as input, we estimate a dense correspondence map to the mesh model for every point on the image grid(see Sec.~\ref{sec:uvnet}). By mapping features from the image grid to the mesh grid according to dense correspondence map, we then recover the 3D coordinates of all the mesh vertices(sec.~\ref{sec:gridmap}) on the mesh grid and finally refine these coordinates by skinning a template mesh model with respect to the recovered mesh vertices(sec.~\ref{sec:refine}).}
    \label{fig:framework}
\end{figure*}

In this section, we introduce our Dual Grid Net (DGN) which is an efficient fully convolutional network architecture for mesh vertex estimation. At its core are consecutive 2D convolutions on two grids -- an image grid and a mesh grid -- where features from one grid can be mapped to another in a differentiable way.

We assume that we are given a canonical hand mesh model which is generic for all users' hands. In a given depth map, every pixel on the hand's surface on the image grid has a correspondence to the mesh surface and estimating this correspondence is equivalent to regressing the pixel's coordinates on the mesh grid (Sec.~\ref{sec:meshmodel}).
 
Starting from a depth map of the segmented hand as input, the associated mesh vertices can be estimated as follows.  First, we estimate a dense correspondence map to the mesh grid for every point in the input point cloud (see Sec.~\ref{sec:uvnet}). We then map features from the image grid to the mesh grid according to dense correspondence map and recover the 3D coordinates of all the mesh vertices(sec.~\ref{sec:gridmap}). We finally refine these coordinates by skinning a template mesh model with respect to the recovered mesh vertices(sec.~\ref{sec:refine}).  This process is illustrated in Figure~\ref{fig:framework}.

\subsection{Mesh model}
\label{sec:meshmodel}
\begin{figure}
    \centering
    \includegraphics[width=\linewidth]{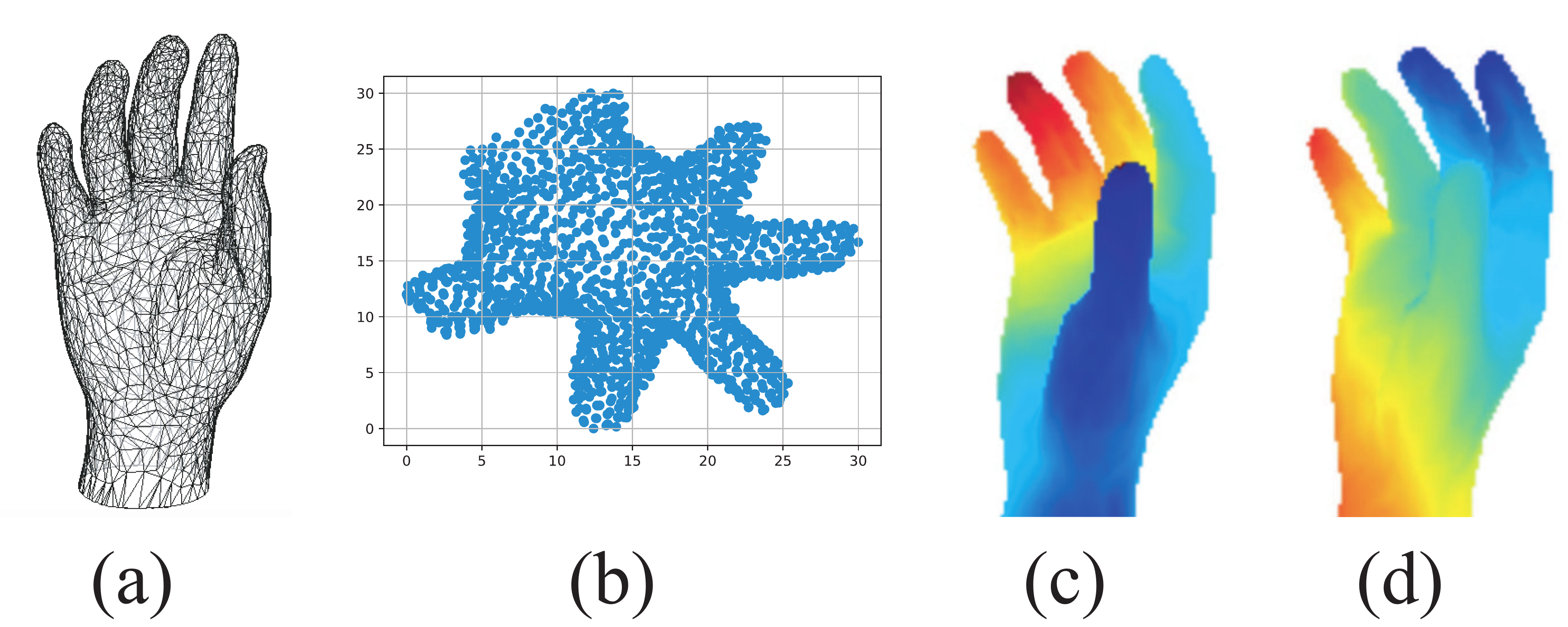}
    \caption{(a) Triangular mesh model used in this work; (b) 2D MDS embedding of the mesh vertices; (c, d) corresponding mesh coordinates on mesh surface.}
    \label{fig:mesh_model}
\end{figure}

We use a triangle mesh model (see Figure~\ref{fig:mesh_model}(a)) with 1721 mesh vertices.
Every point on the mesh surface is associated with a mesh coordinate which depends only on its position on the mesh and is therefore invariant to different hand poses, shapes or view point.  In addition, other properties of points on the mesh surface such as texture, colour or its 3D coordinates in the camera frame can be approximated with linear interpolation of neighbour points on the mesh surface. 

A natural way to parameterize  mesh coordinates is through UV mapping~\cite{uvmapping}, as used in~\cite{alp2017densereg}. However, the mesh unwrapping in UV mapping introduces unnecessary discontinuities along seams. In this work, we use Multidimensional Scaling (MDS)~\cite{Borg1997} instead. For any two points on mesh surface, MDS aims to keep their Cartesian distance \textit{w.r.t.} the mesh coordinates to be as close as possible to the geodesic distance on mesh surface. We set the dimension of mesh coordinates to 2, to allow for 2D convolutions on the mesh grid. The MDS embedding used in this work is shown in Figure~\ref{fig:mesh_model}(b), and the corresponding mesh coordinate on mesh surface in Figure~\ref{fig:mesh_model}(c) and (d) respectively.

\subsection{Mesh Coordinate Estimation}
\label{sec:uvnet}
Similar to~\cite{alp2017densereg}, we start by estimating the 2D mesh coordinates for all pixels from the hand region. We adopt an hourglass network\cite{newell2016stacked} (see Figure~\ref{fig:framework}) as the backbone architecture and apply it in two heads. The first head estimates the 2D mesh coordinates $\mathcal{I}_m$ for all depth pixels while the second head estimates a generic feature map $\mathcal{I}_f$ which will later be mapped to the mesh grid.
Unlike~\cite{joo2018total}, which performs classification followed by residual regression, we adopt a direct regression approach, which we find achieves sufficient accuracy.

Previous works~\cite{ge2019handshapepose,boukhayma20193d, zhang2019end,Malik2018DeepHPSEE} encoded image inputs as a fixed-size latent vector.  Our approach, by using dense mesh coordinates, has two major advantages.  Firstly, it allows us to use a fully convolutional network architecture.  This important difference retains spatial resolution, is more efficient and also translation invariant.  It is also much easier for learning, since supervision at the level of mesh coordinates can be directly placed here.

Secondly, the estimated mesh coordinates establishes a dense correspondence map between captured hand surface to that of mesh. The correspondence map, as we will show in Sec.~\ref{sec:data_term}, allows us to directly embed a lifting energy~\cite{joseph2016fits}, which is beneficial to minimizing the model-fitting error in a self-supervised setting.

\subsection{Mapping from image grid to mesh grid}
\label{sec:gridmap}
\begin{figure}
    \centering
    \includegraphics[width=.9\linewidth]{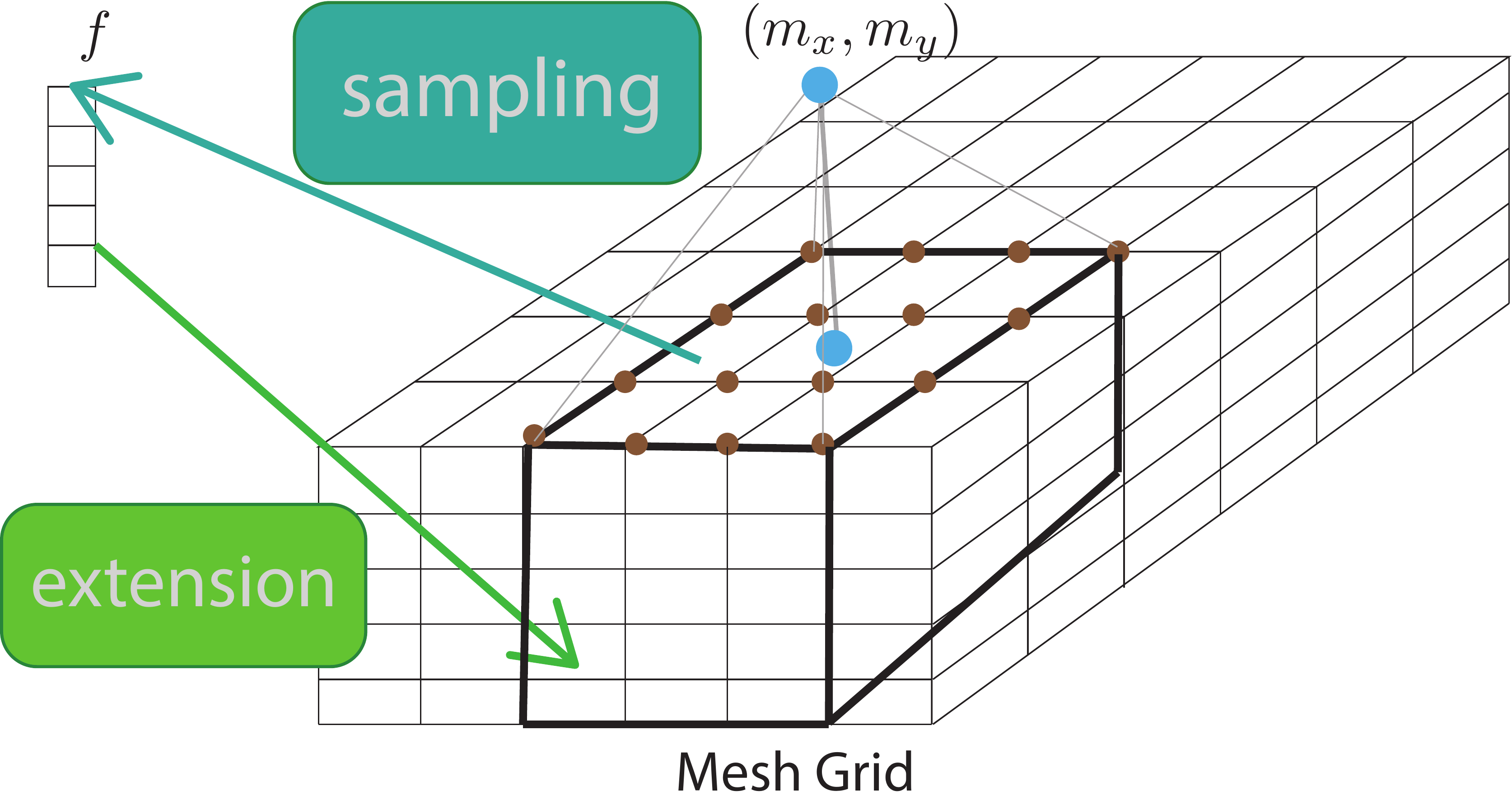}
    \caption{Illustration of extension and sampling process, given the feature to be mapped as $f\in \mathcal{R}^f$ and corresponding coordinate on mesh grid as $(m_x, m_y) \in \mathcal{R}^2$.}
    \label{fig:grid}
\end{figure}

In this section, we show how to recover all mesh vertices, including occluded ones, from the estimated per-pixel mesh coordinate and features on the image grid.
Based on the estimated mesh coordinates, features from a pixel of the hand can be mapped from image grid to mesh grid. 
Similar to~\cite{Matan2018}, we call this process \emph{extension}(see Figure~\ref{fig:grid}). 

More specifically, for any pixel $p$ which belongs to the hand surface, we can regress its coordinate on the mesh grid $m = (m_x, m_y) \in \mathcal{R}^2$ as well as its corresponding feature $f \in \mathcal{R}^d$ as described in previous section.
$f$ is propagated to mesh grid via soft assignment to the neighbours of $m$:
\begin{equation}
    f = \sum_{n\in \Omega(m)} w_n \cdot f.
\end{equation}
$f$ is propagated to the grid point $n$ with a weighting determined by the softmax of its distance to $m$ as follows:
\begin{equation}
    w_n = \frac{e^{-\sigma(n-m)^2}}{\sum_l e^{-\sigma(l-m)^2}},
\end{equation}
where $\sigma = 0.5$. 

We adopt a second hourglass network on the mesh grid, o recover all mesh vertices.
Given that every mesh vertex is associated with a fixed mesh coordinate, the output features of hourglass network is aggregated according to their mesh coordinates of vertices. To this end, we set the number of output feature channels as $3$ and the aggregated feature for each mesh vertices is exactly its estimated 3D coordinates in the camera frame. In turn, this process is named as \emph{sampling} (see Figure~\ref{fig:grid}).

Note that propagated features will only partially occupy the mesh grid due to occlusions.  However, the sampling process requires features from all over the mesh grid. This resembles an  image in-painting process and we leverage the encoder-decoder structure of hourglass to utilize both global and local context when filling in these values.

\subsection{Refining Mesh Vertices}
\label{sec:refine}
We observe that the quality of the rendered mesh by the estimated mesh is sensitive to even small offsets (see Figure~\ref{fig:intro}). At the same time, as we are focusing on a specific model, it is excessive to add any sophisticated network architectures for more accurate mesh vertices estimation.  As an alternative, we propose refining the mesh vertices with a kinematic module without adding learnable parameters.

We refine the estimated mesh vertices by aligning the estimation with a template mesh model and estimating the transformation with a closed form solution. 
More specifically, given the correspondence between estimated vertices $\mathcal{P}_s$ and vertices from the template model $\mathcal{Q}$ for each hand part (palm or finger bone), we estimate a similarity transformation matrix $\mathbf{T}$ by minimizing the Euclidean distance between correspondence points $p_i\!\in\!\mathcal{P}_s$ and $q_i\!\in\!\mathcal{Q}$ as
\begin{equation}
\mathbf{T} = \argmin_{\mathbf{T}} \sum_i\|p_i - \mathbf{T} q_i \|.
\end{equation}
The refined mesh results from posing the template mesh with the similarity transformation matrices through linear blend skinning (LBS). Noticing that $\mathbf{T}$ can also be estimated in closed form with singular value decomposition (SVD).  By using the closed form solution, the refined mesh can be obtained with a single forward pass through the network. Readers may refer to~\cite{sorkine2009least} for more details on estimating the transformation with a closed form solution.

Coordinates of key points can also be obtained from the transformation matrices in a similar way as mesh vertices. Since SVD is differentiable, supervision can be placed on top of the key-point coordinates. As will be shown in Sec.~\ref{sec:exp}, when only given sparse supervision of key-points, our method can accurately recover the mesh.

\subsection{Implementation Details}
\label{sec:implement}
We first segment the hand region with a hourglass network. The input size of image to the hourglass network on the image grid is $64\!\times\!64$ and we set the size of the mesh grid as $16\!\times\!16$. To further reduce computation, we adopt pixel shuffling techniques~\cite{Shi2016RealTimeSI} to decrease the spatial resolution by a factor of 2 on both the image grid and mesh grid. While the number of input and output feature channels are increased by a factor of 4, the number of feature channels in hidden layers remains unchanged. The kernel size of extension and sampling are both set as $8\!\times\!8$.

\section{Self-supervision on unlabelled real data}
\label{sec:self}
Training the network proposed in Section~\ref{sec:network} requires supervision in the form of dense correspondences and vertex locations which is impossible to annotate for real world data. While the network can be initialized with synthetic depth maps, as shown in the experiments, the large domain gap between real and synthesized depth map gives rise to compromised accuracy. On the other hand, since the network also serves as a differentiable renderer, the natural question that arises is whether or not we can include a model-fitting loss term into the training loss for self-supervised learning.

Similar to the conventional model fitting energy, the self-supervision term is formulated as follows,

\begin{equation}
L(\theta) = L_{\text{data}}(\theta) + \lambda_1 L_{\text{prior}}(\theta) + \lambda_2 L_{\text{mv}}(\theta).
\end{equation}

\noindent This data fitting loss is similar to conventional model-fitting energy terms. It is  composed of a data term $L_\text{data}$, which measures how the rendered depth map resembles the input depth map; kinematic priors $L_\text{prior}$ which constrain the estimate to be kinematically feasible and a multi-view consistency term $L_\text{mv}$ which can be used in calibrated multi-camera setups to handle self-occlusion.

\subsection{Data Terms}
\label{sec:data_term}
The data term is composed of an ICP term and a lifting energy term:

\begin{equation}
L_{\text{data}}(\theta) = L_{\text{ICP}}(\theta) + \alpha L_{\text{lifting}}(\theta).
\end{equation}

Similar to~\cite{taylor2017articulated}, we consider only a data-to-model term, \ie only minimizing the distance between every depth point to its correspondence on the mesh surface. Ignoring the model-to-data term makes the loss robust to occlusions which is useful for hand-object or hand-hand interactions.

The \textbf{ICP term} measures the disparity between points to its projection on the mesh surface as follows,
\begin{equation}
L_{\text{ICP}}(\theta) = \sum_{i\in \mathcal{I}} \min_{j\in M(\theta)} d(i, j),
\end{equation}
where the projection on estimated mesh surface $M(\theta)$ is approximated by finding the nearest vertices from mesh model based on the distance function $d$. We use smooth L1 loss function as $d(\cdot, \cdot)$.
Similar to~\cite{tagliasacchi2015robust}, we restrict points only to find correspondences in the frontal surface of the mesh. 

In addition, we leverage the correspondence map and minimize the distance between points to their estimated correspondences on the mesh surface via a \textbf{lifting term}: 
\begin{equation}
L_{\text{lifting}}(\theta) = \sum_{i\in \mathcal{I}} d(i, f(i | \theta)).
\end{equation}
where $f(i | \theta)$ estimates the 3D coordinates of correspondence of $i$ on the mesh surface, given the estimated mesh coordinate of $i$ through the sampling process(see Figure~\ref{fig:grid}). The lifting term simultaneously optimizes over the correspondence map $\mathcal{I}_m$ on the image grid and the coordinate map $\mathcal{J}_o$ on the mesh grid (see Figure~\ref{fig:framework}). As such, this helps a more efficient gradient flow to different network stages.

\subsection{Kinematic Priors}
\label{sec:prior_term}
The kinematic prior terms are defined as
\begin{equation}
L_{\text{prior}}(\theta) = L_{\text{collision}}(\theta) + \kappa_1 L_{\text{arap}} + \kappa_2 L_{\text{offset}}(\theta).
\end{equation}
\noindent The \textbf{collision term} $L_{\text{collision}\theta}$ penalizes collisions between any pair of joints as follows:
\begin{equation}
    L_{\textbf{collision}}(\theta) = \sum_{i, j} \max(t - \|p_i - p_j \|, 0),
\end{equation}
where $p_i$ and $p_j$ are the 3D coordinate of the corresponding joints. We set the threshold $t = 5\text{mm}$ for all pair of joints.

The \textbf{as rigid as possible term} $L_{\text{arap}}(\theta)$  constrains the local deformation of estimated mesh surfaces to be rigid, similar to~\cite{Sorkine:2007:ASM:1281991.1282006}. 
\begin{equation}
    L_{\textbf{arap}} = \|\mathcal{P}_s - \mathcal{P}_s\|^2
\end{equation}
where $\mathcal{P}_s$ is the original mesh vertices estimation and $\mathcal{P}_r$ is the refined one through linear blend skinning, which is guaranteed to be rigid for each part.

\begin{figure}
    \centering
    \includegraphics[width=\linewidth]{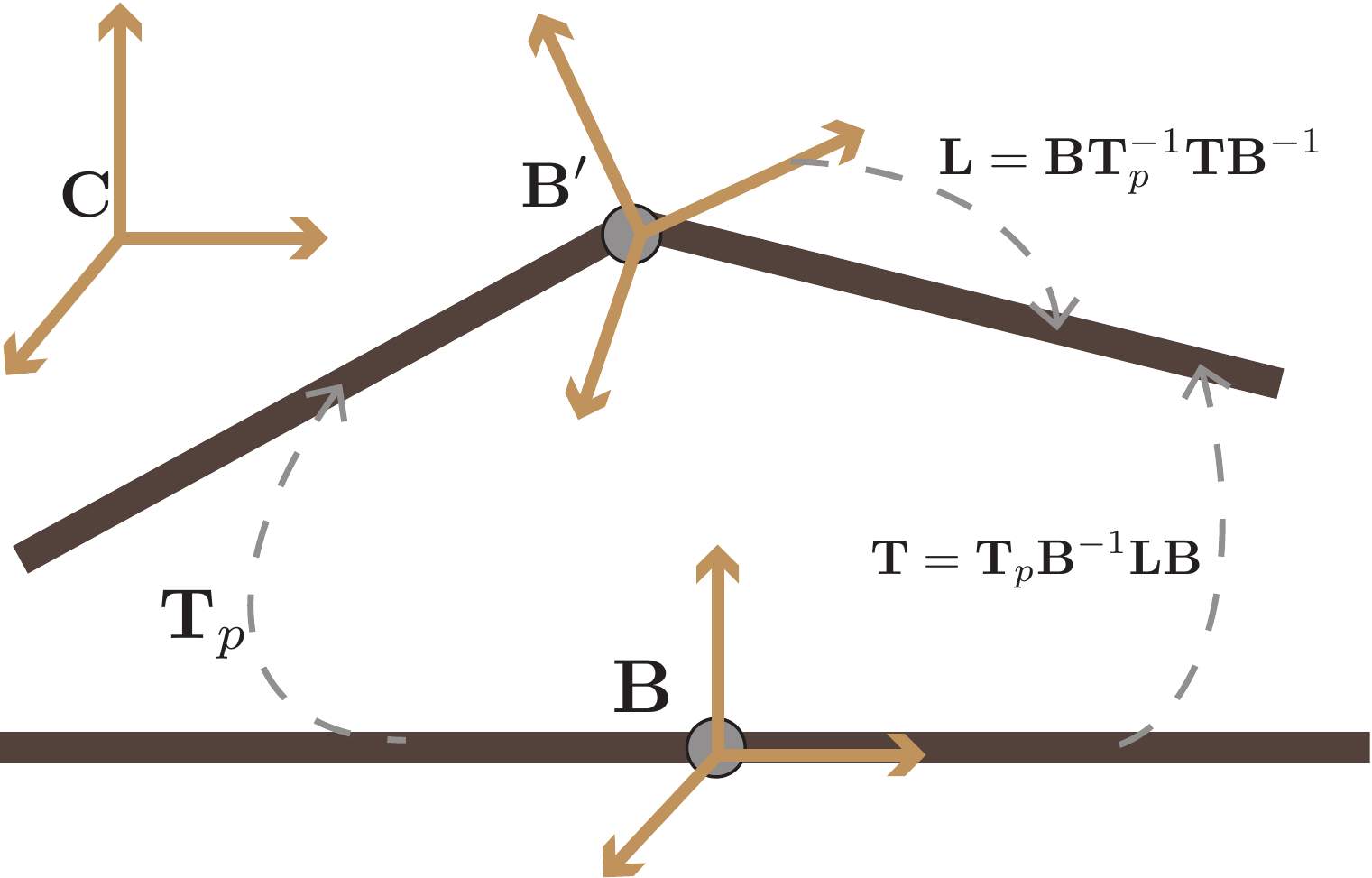}
    \caption{Illustration of the relation ship between local transformation $\mathbf{L}$ with respect to the local bone frame $\mathbf{B}$ and global transformation $\mathbf{T}$ with respect to the camera frame $\mathbf{C}$.}
    \label{fig:trans}
\end{figure}

In section~\ref{sec:refine}, we show how to estimate the similarity transformation $\mathbf{T}$(see Figure~\ref{fig:trans}) with respect to the camera frame for each hand part. In other words, $\mathbf{T}$ transforms the bone from a neutral pose to the current one with respect to the camera frame. From the perspective of forward kinematics, $\mathbf{T}$ is generated as follows,
\begin{equation}
\mathbf{T} = \mathbf{T}_p \cdot \mathbf{B}^{-1} \cdot \mathbf{L} \cdot \mathbf{B},
\end{equation}

\noindent where $\mathbf{T}_p$ is the parent transformation matrix, $\mathbf{B}$ is the bone frame in the neutral pose with which z-axis is aligned with its parent bone, the origin is placed at the joint. $\mathbf{L}$ is the rotation matrix with respect to the bone frame $\mathbf{B}$.
Since $\mathbf{B}$ is given in the original mesh model and $\mathbf{T}_p$ is known from previous estimation, $\mathbf{L}$ can be recovered with a closed form solution.

We rewrite the local transformation matrix $\mathbf{L}$ as $[\mathbf{S}\mathbf{R} | t]$, where $\mathbf{S}\in R^{3\times3}$ is a diagonal matrix scaling the matrix, $\mathbf{R}\in R^{3\times3}$ is the rotation matrix, $t\in R^3$ is the translation. 
Notice that besides the wrist, there is no translation on the rest finger joints. We thus penalize translations in the finger's local transformation with an \textbf{offset term}
\begin{equation}
    L_{\text{offset}} = \sum_{i\in \mathcal{F}} \|t_i\|^2,
\end{equation}
where $\mathcal{F}$ represents all the finger joints.

We don't add further push constraints over the joint angles since synthesized data with supervision is also fed to the network to regularize the estimation. Given that joint angles can be calculated from local transformation $\mathbf{L}$ with a closed form solution, joint angle constraints can be easily added if necessary. 

\subsection{Multiple view consistency}
\label{sec:mv_term}
To handle severe self-occlusions and missing inputs due to holes in noisy depth inputs, we further add multi-view consistency constraints for real data captured on a multi-camera rig: 

\begin{equation}
L_{\text{mv}}(\theta) = L_{\text{vertex}}(\theta) + \eta_1 L_{\text{ICP}}(\theta) + \eta_2 L_{\text{lifting}}(\theta).
\end{equation}
By calibrating the extrinsics of the camera, the \textbf{vertex term} $L_\text{vertex}$ minimizes the distance between mesh vertices to their robust average (median in this paper) in the canonical frame.
The \textbf{ICP term} $L_\text{ICP}$ and \textbf{lifting term} $L_\text{lifting}$ works similarly to the aforementioned single view cases, with the only difference that estimated mesh model is mapped to another camera frame and matched against the corresponding depth map.

\subsection{Active data augmentation by estimation}
\label{sec:aug}
Since the proposed method could recover the hand mesh, we propose a strategy to actively feed synthesized data given the estimated mesh on real data to the network. The supervision from the synthesized data provides more realistic poses and helps the network to better recover from wrong estimation.
According to experiments, we find this strategy to be useful to stabilize the self-supervision training and further decrease the model fitting error on unlabelled training data.

\section{Experimentation}
\label{sec:exp}
\subsection{Dataset and evaluation protocols}
We evaluate our method on the NYU Hand Pose Dataset\cite{tompson2014real}.  It currently the only publicly available multi-view depth dataset and features sequences captured by 3 calibrated and synchronized PrimeSense depth cameras. It consists of $72757 \times 3$ frames for training and $8252 \times 3$ for testing. NYU is highly challenging as the depth maps are noisy and the sequences cover a wide range of hand poses. In addition, we synthesize a dataset of 20K depth maps of various hand poses with random holes and depth noise to evaluate the trained network's ability to generalize to new synthesized samples. Our method is highly efficient and achieves 63.1 FPS on an Nvidia 1080Ti GPU.

Following the protocol of~\cite{tompson2014real} and previous works, we quantitatively evaluate a subset of 14 joints with two standard metrics: mean joint position error (in mm) averaged over all joints and frames, and the percentage of success frames, \ie, frames with all predictions are within a certain threshold~\cite{taylor2012vitruvian}. Readers may refer to supplementary materials and video for qualitative results.

\subsection{Training with only synthesized data}
We first evaluate how a network trained on synthesized data can generalize to newly synthesized data and real data. The synthesized data is rendered from a mesh model with various poses and shapes and then corrupted with random depth noise and holes. Data is synthesized in an on-line manner and around 7.2 million synthesized samples are fed into the network for training. Table~\ref{tab:self-compare} (synt(test on synt), synt(mesh vertices)(test on synt), synt(refined mesh vertices)(test on synt)) shows that the proposed kinematic module successfully reduces the average error over all mesh vertices from 14.75mm to 7.65mm. The network can also generalize to newly synthesized samples and achieves a high accuracy with only 7.1mm mean joint position error. However, the accuracy deteriorates dramatically when testing on real-world depth maps. The mean average joint error increases almost three-fold to 23.21mm. This shows that even though it encounters data augmented with random noise, the network readily over-fits to the rasterization artifacts and hand shapes of synthesized depth maps.

\subsection{Ablation studies}
\textbf{Variations in training data.} We investigate how different training data and different supervision impacts the accuracy. First, we train only with the $8252\times3$ testing samples to check how well self-supervision can fit the mesh model to depth maps. We then trained with all training data, but in a single view setting to check how a multi-view set up impacts performance. Finally, we also look into supervision with sparse key-points to check if the proposed network accurately recover the mesh vertices and the key-points on unseen samples in testing set.

Interestingly, training directly on the test samples gives rise to a higher mean joint position error than when training on a larger training set that excludes the test samples(14.50mm vs 13.09mm, see Table~\ref{tab:self-compare}). We attribute this to the poor initialization of the network when trained on synthesized data and the possibility of getting trapped in local minima since first order based optimization is used during back-propagation.
However, if the amount of training data increases, mean joint position error decreases. This justifies the benefits of data-driven approaches over conventional model-based trackers which optimizes each frame independently.

\begin{figure}
    \centering
    \includegraphics[width=.9\linewidth]{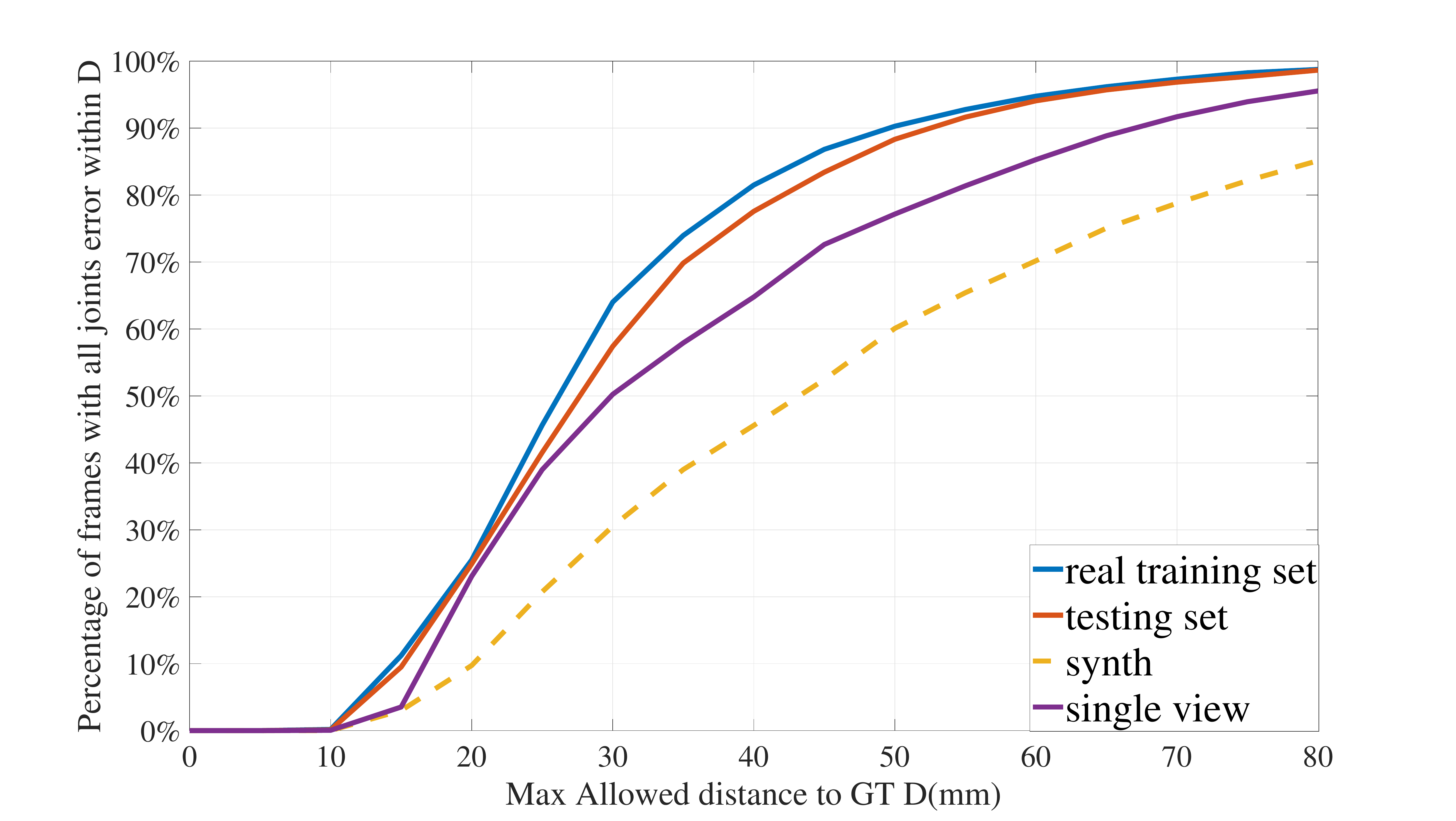}
    \caption{Impact of using different dataset for  self-supervision.}
    \label{fig:ablation}
\end{figure}

As shown in Figure~\ref{fig:intro} (see more qualitative examples in the supplementary materials), our method can accurately reconstruct the 3D mesh model given only sparse key-point supervision. When it comes to mean joint position error, the estimation is highly accurate with only 8.5 mean joint position error (see Table~\ref{tab:self-compare}).  Furthermore, 67.8\% of frames have a maximum error below 20mm and 85.3\% below 30mm respectively (see Table~\ref{fig:ablation}).
 
\textbf{Impact of self-supervision loss terms.} We study the individual contributions of the different self-supervision loss terms by training without the $L_{\text{lifting}}$, $L_{\text{collision}}$, $L_{\text{arap}}$, $L_{\text{offset}}$ and active augmentation techniques. 
According to Table~\ref{tab:self-compare} and Figure~\ref{fig:ablation}, without the lifting energy techniques, the average error increases by 1.41mm from 13.09mm to 14.50mm.  The 
percentage of successful frames drops by 7\% from 64\% to 57\% on the error threshold of 30mm. This validates the benefits of the lifting energy. The contributions of the other terms can also be validated as we observe similar decreases in accuracy when they are omitted.

\subsection{Comparison to state-of-the-art}
\begin{figure}
    \centering
    \includegraphics[width=.9\linewidth]{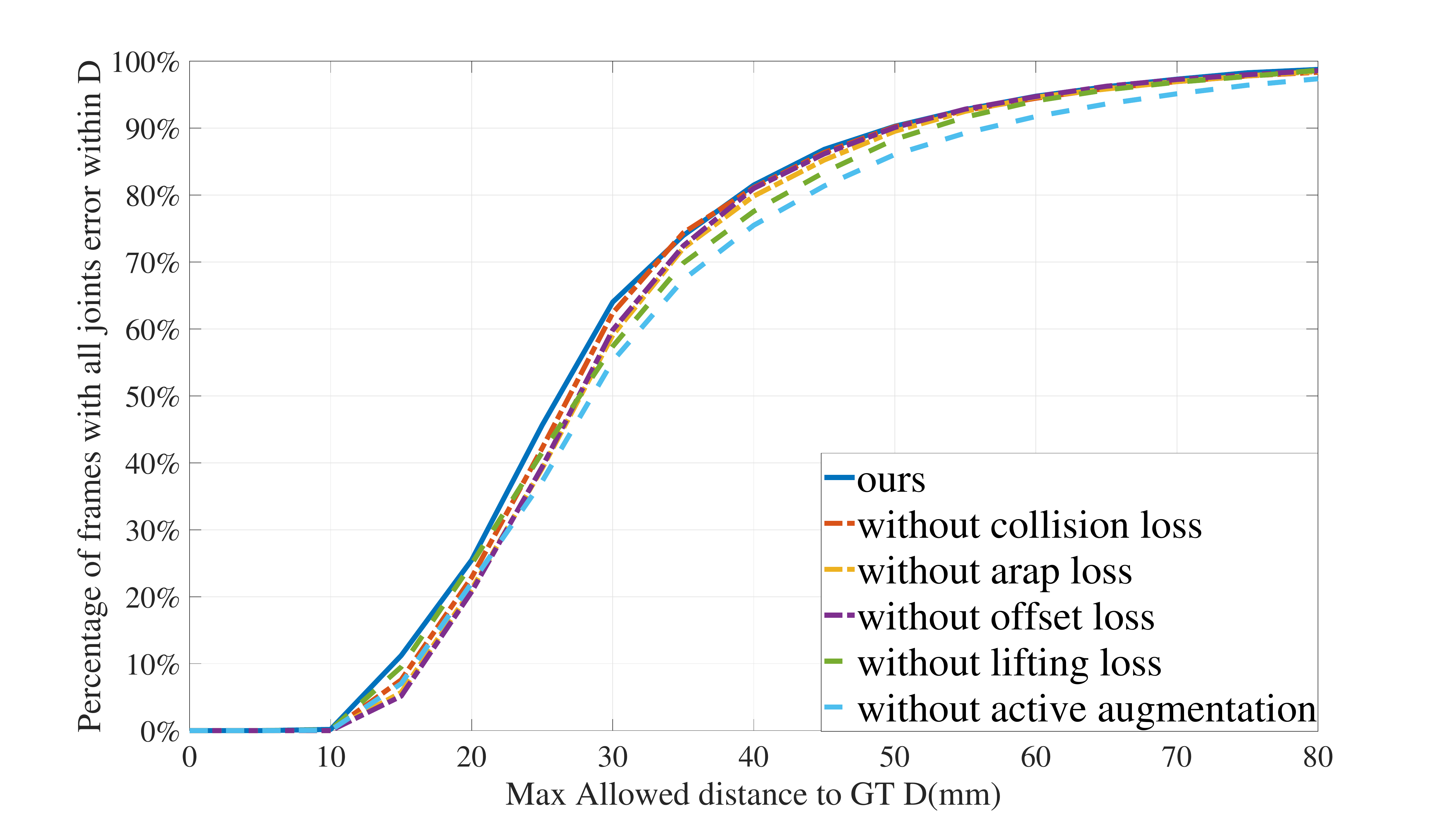}
    \caption{Impact of different loss terms and active data augmentation on self-supervised learning.}
    \label{fig:soa}
\end{figure}

We compare our results to the most recently proposed state-of-arts \cite{rad2018feature,moon2018v2v,ge2018point,chen2018shpr,poier2019murauer,wan2017dense,chen2017pose,oberweger2017deepprior++,guo2017region,ge20173d,Malik2018DeepHPSEE,xu2017lie,wan2017crossing,oberweger2015training,zhou2016model}. As shown in Table~\ref{tab:state-of-art}, when trained with keypoint annotations, our method outperforms all other state-of-arts except~\cite{moon2018v2v} and~\cite{rad2018feature} with respect to mean joint position error.  In addition, according to Figure~\ref{fig:soa}, our method performs similarly to~\cite{ge2018point,poier2019murauer} when the error threshold is larger than 10mm and outperforms all other methods except~\cite{rad2018feature}.  We note however that~\cite{moon2018v2v} report an ensemble prediction result.  This is impractical for real time use; in comparison, our method is highly efficient and runs at 63.1 FPS on an NVidia 1080Ti GPU. Furthermore, our method out-performs~\cite{moon2018v2v} when compared its single model result. The work of~\cite{rad2018feature} leverages domain adaption techniques to better utilize synthesized data. This is complimentary to our proposed method and we look forward to incorporating this in our future work. It is also worth noting that key-point estimation is a byproduct of our proposed method, which has the primary aim of recovering the mesh vertices.

We also compare our self-supervision method with~\cite{dibra2017refine}, which to best of our knowledge is the only other unsupervised method. As is shown in Figure~\ref{fig:soa}, our network outperforms~\cite{dibra2017refine} by a large margin for the percentage of successful frames at error thresholds higher than 25mm. We achieve a higher accuracy for two reasons. First, our mesh parameterization allows the method to be robust to small estimation offsets while~\cite{dibra2017refine} uses joint angles, which tend to propagate errors from parent joints to children joints.
Second, there are no gradients in their \emph{depth term}(Eq. 6 in~\cite{dibra2017refine}) associated with unexplained points from the depth map which we handle with our proposed data term. 

We further compare our self-supervision method with fully supervised deep learning methods. Surprisingly, when trained without any human label, our self-supervision based method achieves competitive results and even out-performs several fully supervised methods\cite{guo2017region,ge20173d,Malik2018DeepHPSEE,xu2017lie,wan2017crossing,oberweger2015training,zhou2016model}. This highly encouraging results suggests that our method can be applied to provide labels for RGB datasets with weak supervision from depth maps.

\begin{table}[htbp]
    \centering
    \begin{tabular}{l c}
    \hline
        Method &  Mean joint error \\ \hline
        ours (fully supervised) & 8.5mm\\
        ours (self-supervised)& 13.09mm\\ \hline
        synt(test on synt) & 7.10mm\\
        synt(mesh vertices) (test on synt) & 14.75mm\\
        synt(refined mesh vertices) (test on synt) & 7.65mm\\
        synt(test on real) & 23.21mm\\ \hline\hline
        train on test& 14.50mm\\ 
        single view & 16.96mm\\  \hline\hline
        without active augmentation & 14.52mm\\
        without $L_{\text{lifting}}$ & 14.50mm\\
        without $L_{\text{collision}}$ & 13.85mm\\
        without $L_{\text{arap}}$ & 14.06mm\\
        without $L_{\text{offset}}$& 14.12mm\\ 
    \end{tabular}
    \caption{\textbf{Ablation study and self comparison.} We report mean joint error averaged over all joints and frames.}
    \label{tab:self-compare}
\end{table}

\begin{figure}
    \centering
    \includegraphics[width=.9\linewidth]{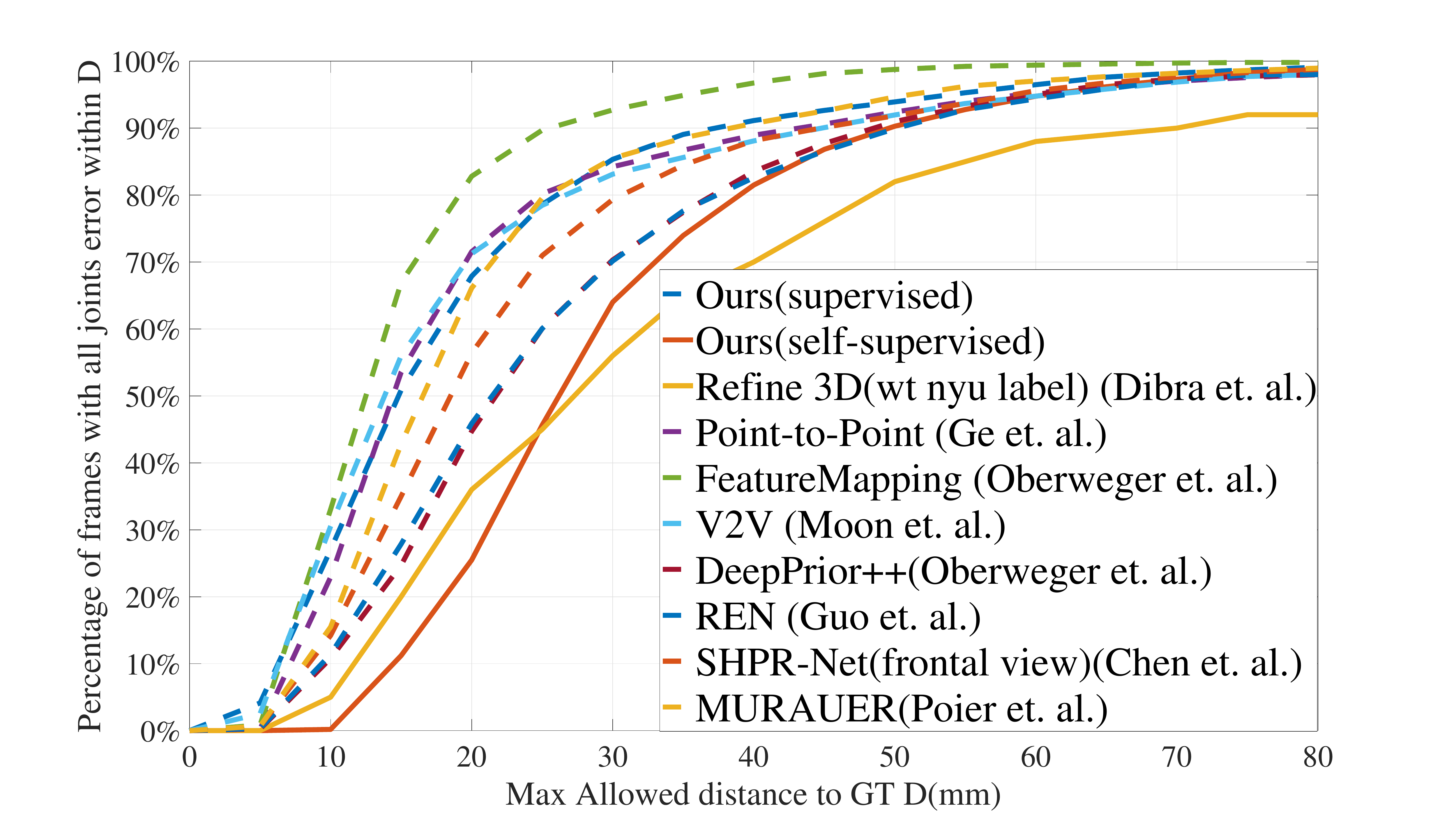}
    \caption{Comparison to fully supervised (dashed line) and self-supervised (solid line) state-of-arts.}
    \label{fig:my_label}
\end{figure}

\begin{table}[htbp]
    \centering
    \begin{tabular}{l c}
    \hline
        Method &  mean joint error \\ \hline
        ours (supervised) & 8.5mm\\ 
        ours (self-supervised) & 13.1mm\\ \hline
        FeatureMapping\cite{rad2018feature} & 7.4mm\\
        V2V(ensemble)\cite{moon2018v2v} & 8.4mm\\
        V2V(single model)\cite{moon2018v2v} & 9.2mm\\
        Point-to-Point\cite{ge2018point} & 9.0mm\\
        SHPR(three views)\cite{chen2018shpr} & 9.4mm\\
        MURAUER\cite{poier2019murauer} & 9.5mm\\
        DenseReg\cite{wan2017dense} & 10.2mm\\
        Pose-REN\cite{chen2017pose} & 11.8mm\\
        DeepPrior++\cite{oberweger2017deepprior++} & 12.2mm\\
        REN-4x6x6\cite{guo2017region} & 13.4mm\\
        3DCNN\cite{ge20173d} & 14.1mm\\
        DeepHPS(fine-tuned)\cite{Malik2018DeepHPSEE} & 14.2mm\\
        Lie-X\cite{xu2017lie} & 14.5mm\\
        CrossingNet\cite{wan2017crossing} & 15.5mm\\
        Feedback\cite{oberweger2015training} & 15.9mm\\
        DeepModel\cite{zhou2016model} & 17.0mm\\
    \end{tabular}
    \caption{\textbf{Comparison with fully supervised state-of-the-art.} We report mean joint error averaged over all joints and frames. All methods are tested on the NYU\cite{tompson2014real} test set.}
    \label{tab:state-of-art}
\end{table}

\section{Conclusion}
We propose a new network architecture to regress thousands of mesh vertices from single depth map with efficient fully convolutional network on 2D grids. 
We show that when initialized with synthesized data, the network could accurately recover the hand mesh vertices with only sparse key point supervision. When given only unlabeled real world dataset, the proposed network can be fine tuned in a self-supervision manner and provide comparable accuracy to state-of-arts with multi-camera rig during training.  Finally, although this paper focuses on depth map input the human hand, since we use 2D FCN, the proposed method can be readily applied to RGB images without any changes to the architecture, when RGB annotation is available.
{\small
	\bibliographystyle{ieee}
	\bibliography{egbib}
}

\pagebreak
\begin{center}
	\textbf{\large Supplemental Materials}
\end{center}
\section{Qualitative results}
\begin{figure*}
	\centering
	\includegraphics[width=1.05\linewidth]{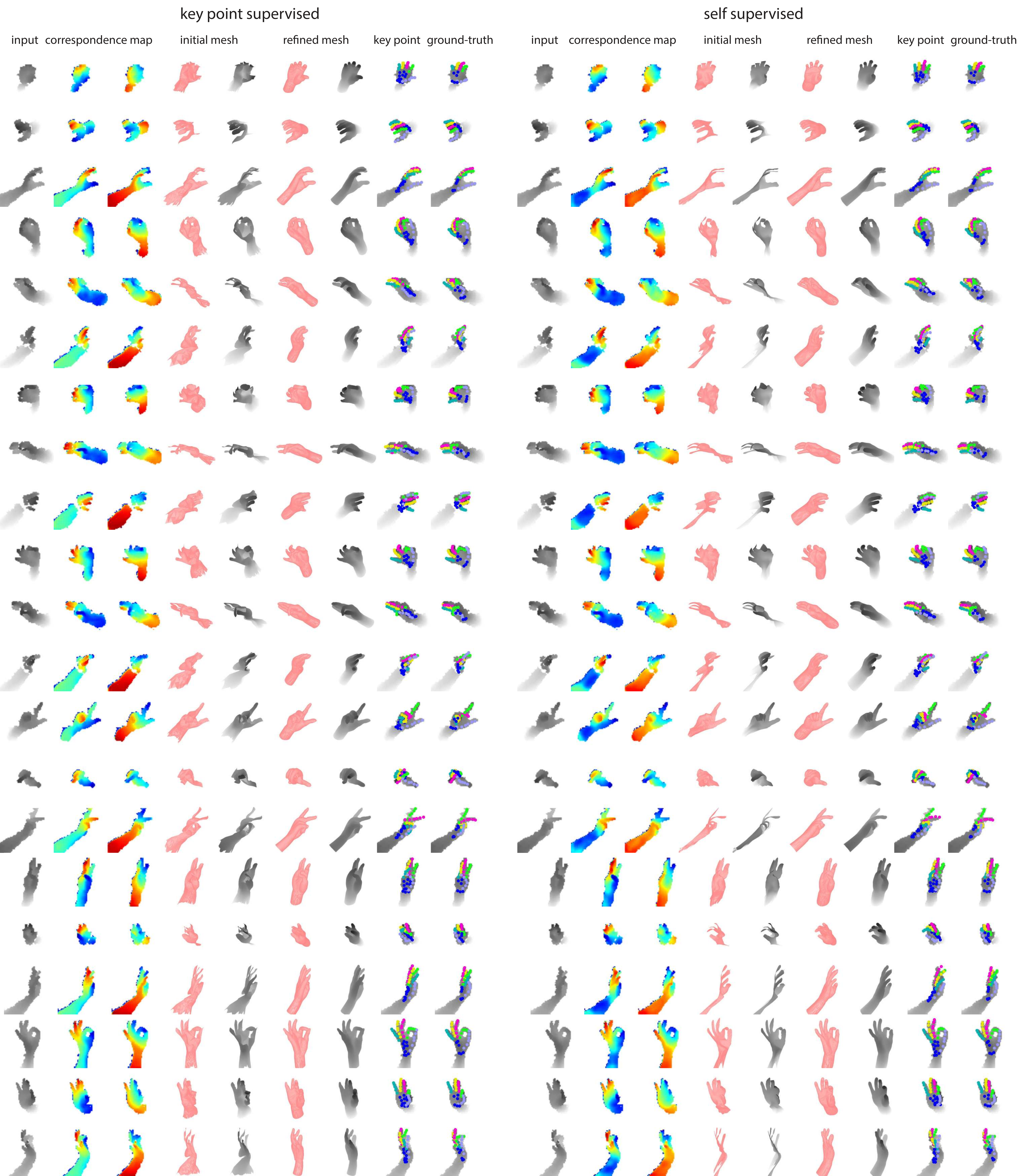}
	\caption{Qualitative results on NYU dataset. We visualize the correspondence map with each mesh coordinate, the rendered shading and depth map of the initial estimated mesh model and refined ones, as well as estimated and ground truth key-point.}
	\label{fig:qua1}
\end{figure*}

\begin{figure*}
	\centering
	\includegraphics[width=1.05\linewidth]{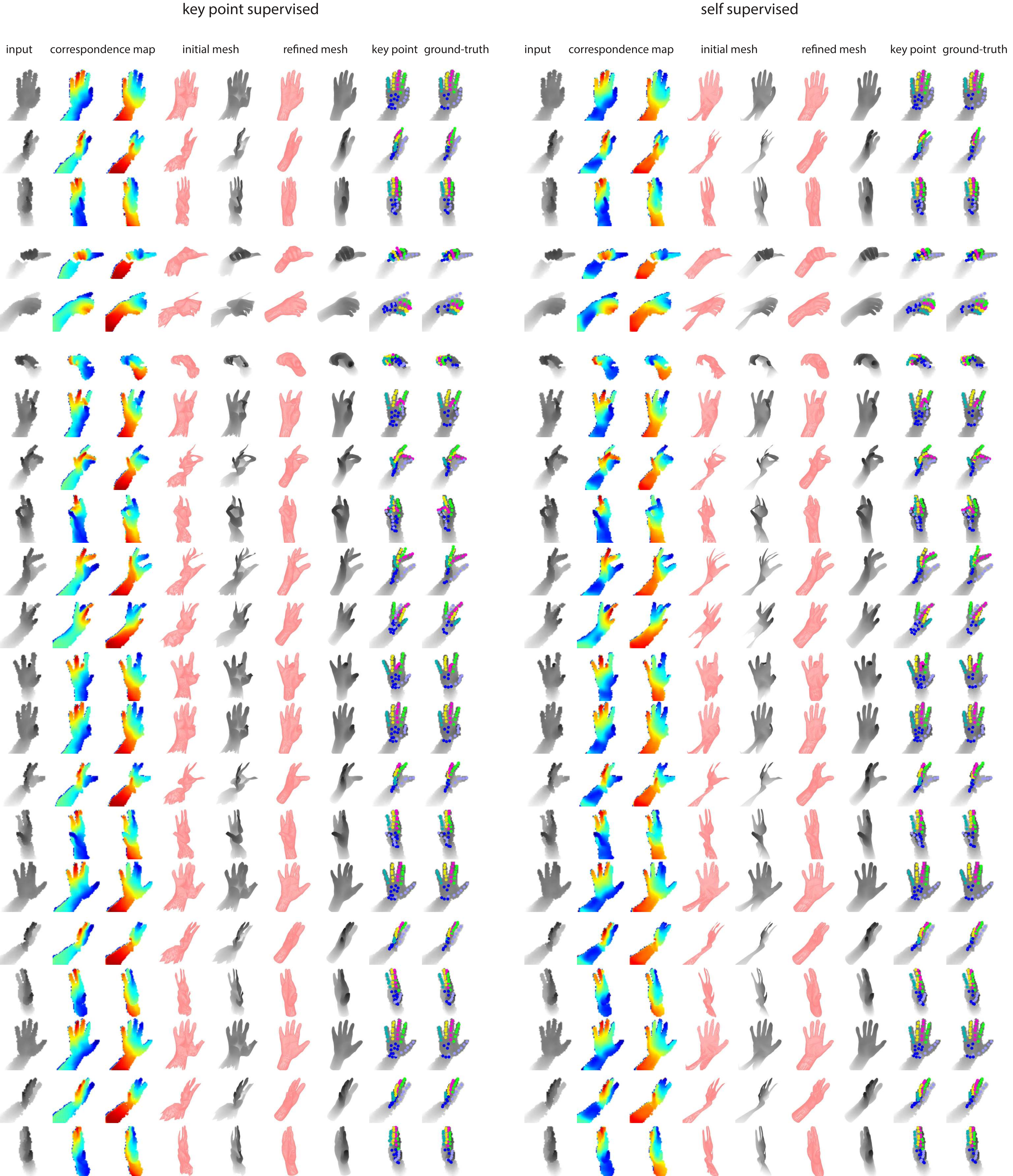}
	\caption{Qualitative results on NYU dataset. We visualize the correspondence map with each mesh coordinate, the rendered shading and depth map of the initial estimated mesh model and refined ones, as well as estimated and ground truth key-point.}
	\label{fig:qua2}
\end{figure*}

\begin{figure*}
	\centering
	\includegraphics[width=1.05\linewidth]{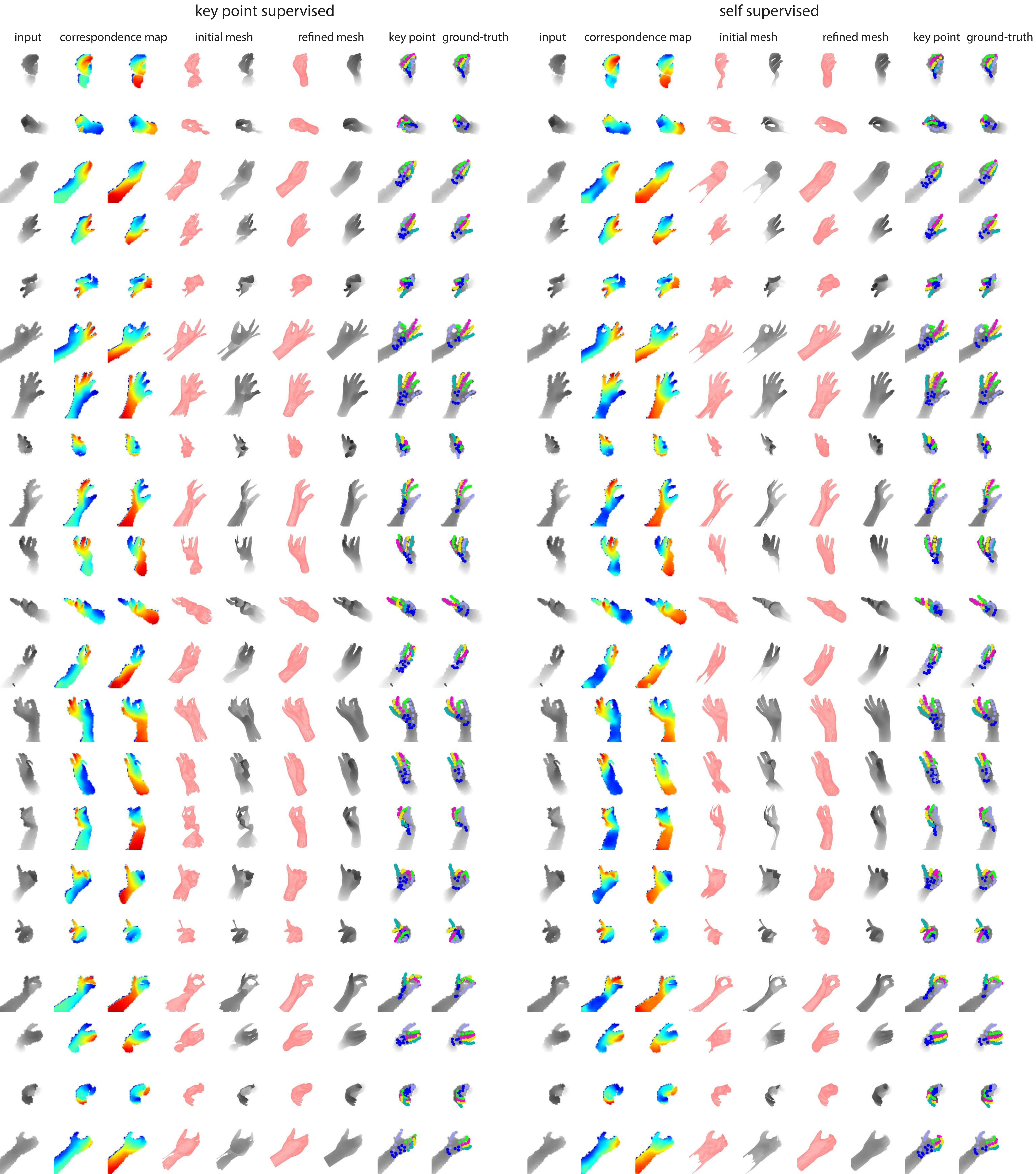}
	\caption{Qualitative results on NYU dataset. We visualize the correspondence map with each mesh coordinate, the rendered shading and depth map of the initial estimated mesh model and refined ones, as well as estimated and ground truth key-point.}
	\label{fig:qua3}
\end{figure*}

We show more qualitative results on the testing set of NYU dataset in Fig.~\ref{fig:qua1}, \ref{fig:qua2} and~\ref{fig:qua3}.
Left column shows results trained by the sparse key point supervision. Right column shows results trained by the proposed self-supervision method.
Readers may also refer to the attached video to check qualitative results on more frames.

\section{Self-supervision training error}
\begin{figure}
	\centering
	\includegraphics[width=\linewidth]{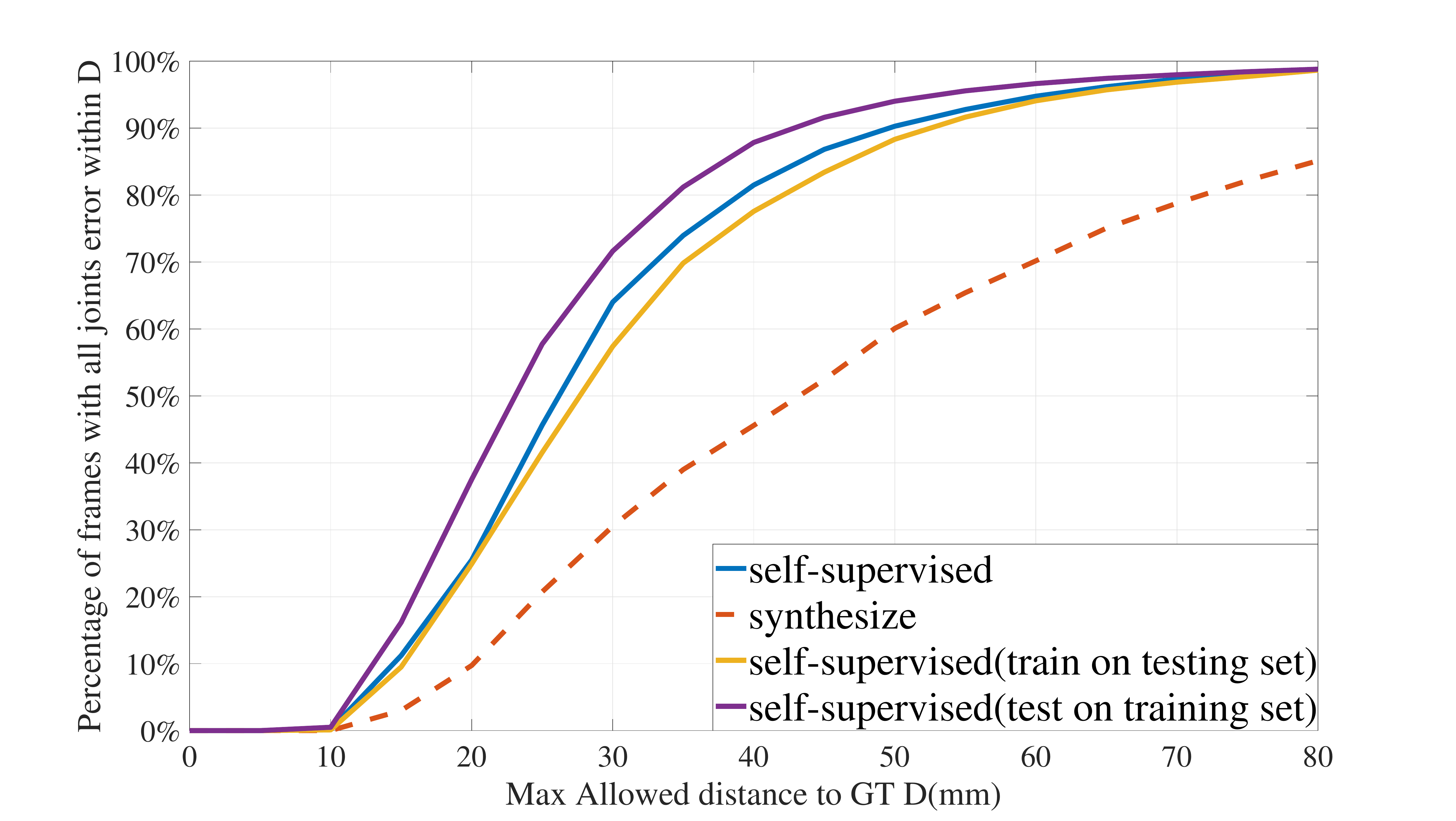}
	\caption{Comparison of using different dataset for training and testing for proposed self-supervision method.}
	\label{fig:res}
\end{figure}

We investigate how well the proposed self-supervision method can fit to the training set itself, \ie, the training error, as ``self-supervised(test on training set)'' in Tab.~\ref{tab:self-compare} and Fig.~\ref{fig:res}. Since our self-supervision method can be potentially applied for automatic annotation of depth frames and accompanied RGBs, its training error indicates how accurate can the annotation be.

We compare the training error against its corresponding testing error, \ie, on the unseen testing set with the same network (``self-supervised''), as well as the testing error trained only with synthesized dataset (``synthesize''), and training error on the testing set (``self-supervised(train on testing set)''), which is roughly $9$ times smaller than the training set.
As expected, compared to accuracy on the testing set, the mean joint error on training set decreases by $1.2$mm from $13.1$mm to $11.9$mm according to Tab.~\ref{tab:self-compare}, and around 10\% more successful frames on the error threshold between $20$ to $40$mm according to Fig.~\ref{fig:res}. 
When comparing with the recent proposed state-of-arts with complicated network architectures and trained with full supervision\cite{xu2017lie,ge20173d,guo2017region,oberweger2017deepprior++,chen2017pose,wan2017dense}, our self-supervision method provides with competitive or even higher accuracy.
This validates our self-supervision method can provide with highly accurate annotation. 

In addition, as already discussed in the paper, the accuracy of self-supervision method is also impacted by the size of the dataset, even when no human label is provided. This infers that accuracy can be further improved when collecting a larger scale dataset.

\begin{table}[htbp]
	\centering
	\begin{tabular}{l c}
		\hline
		Method &  Mean joint error \\ \hline
		self-supervised & 13.1mm\\
		synthesize & 23.2mm\\ 
		self-supervised(test on training set) & 11.9mm\\
		self-supervised(train on testing set)& 14.5mm\\ \hline \hline
		Lie-X\cite{xu2017lie} & 14.5mm\\
		3DCNN\cite{ge20173d} & 14.1mm\\
		REN-4x6x6\cite{guo2017region} & 13.4mm\\
		DeepPrior++\cite{oberweger2017deepprior++} & 12.2mm\\
		Pose-REN\cite{chen2017pose} & 11.8mm\\
		DenseReg\cite{wan2017dense} & 10.2mm\\
	\end{tabular}
	\caption{\textbf{Ablation study and self comparison.} We report mean joint error averaged over all joints and frames.}
	\label{tab:self-compare}
\end{table}

\newpage

\newpage

\newpage

\end{document}